\documentclass[lettersize,journal]{IEEEtran}
\usepackage{amsmath,amsfonts}
\usepackage{algorithmic}
\usepackage{algorithm}
\usepackage{array}
\usepackage[caption=false,font=normalsize,labelfont=sf,textfont=sf]{subfig}
\usepackage{textcomp}
\usepackage{stfloats}
\usepackage{url}
\usepackage{verbatim}
\usepackage{graphicx}
\usepackage{cite}
\usepackage{bbm}
\usepackage{comment}
\usepackage{booktabs}
\newcommand{\tableCellHeight}{1}
\newcommand{\tabstyle}[1]{
  \setlength{\tabcolsep}{#1}
  \renewcommand{\arraystretch}{\tableCellHeight}
  \centering
}

\usepackage{xcolor}

\usepackage{float}

\begin{document}

\title{Dynamic Instance Domain Adaptation}

\author{Zhongying Deng, Kaiyang Zhou, Da Li, Junjun He, Yi-Zhe Song, Tao Xiang
\thanks{Zhongying Deng (e-mail: z.deng@surrey.ac.uk), Yi-Zhe Song and Tao Xiang are with the Centre
for Vision Speech and Signal Processing (CVSSP), University of Surrey, and also with iFlyTek-Surrey Joint Research Center on Artificial Intelligence, University of Surrey, Guildford GU2 7XH, United Kingdom.}
\thanks{Kaiyang Zhou is with Nanyang Technological University, Singapore 639798.}
\thanks{Da Li is with Samsung AI center, Cambridge CB1 2JH, United Kingdom.}
\thanks{Junjun He is with Shanghai Jiao Tong University, Shanghai 200240, China.}
}

\maketitle

\begin{abstract}
Most existing studies on unsupervised domain adaptation (UDA) assume that each domain's training samples come with domain labels (e.g., painting, photo). Samples from each domain are assumed to follow the same distribution and the domain labels are exploited to learn domain-invariant features via feature alignment. However, such an assumption often does not hold true---there often exist numerous finer-grained domains (e.g., dozens of modern painting styles have been developed, each differing dramatically from those of the classic styles).  Therefore, forcing feature distribution alignment across each artificially-defined and coarse-grained domain can be ineffective.  In this paper, we address both single-source and multi-source UDA from a completely different perspective, which is to \emph{view each instance as a fine domain}. Feature alignment across domains is thus redundant. Instead, we propose to perform \emph{dynamic instance domain adaptation} (DIDA). Concretely, a dynamic neural network with adaptive convolutional kernels is developed to generate instance-adaptive residuals to adapt domain-agnostic deep features to each individual instance. This enables a shared classifier to be applied to both source and target domain data without relying on any domain annotation. Further, instead of imposing intricate feature alignment losses, we adopt a simple semi-supervised learning paradigm using only a cross-entropy loss for both labeled source and pseudo labeled target data. 
Our model, dubbed DIDA-Net, achieves state-of-the-art performance on several commonly used single-source and multi-source UDA datasets including Digits, Office-Home, DomainNet, Digit-Five, and PACS.
\end{abstract}


\section{Introduction}
\label{sec:intro}

Deep convolutional neural network (CNN) based models have proven to be highly effective to computer vision tasks such as image classification, provided that a sufficiently large training dataset is available and the test data follow the same distribution as that of the training set~\cite{he2016deep}. However, in real-world deployments, the performance of these models on an unseen test set often drops significantly compared to the validation performance on the source data. This is a common issue that is caused by distribution shift (also known as domain shift) between the source and target data~\cite{ben2010theory}, which violates the i.i.d.~assumption made by most learning algorithms.
Unsupervised domain adaptation (UDA) has been extensively researched to solve the domain shift problem~\cite{gretton2012kernel,long2015learning,long2016unsupervised,hou2016unsupervised,tzeng2014deep,bhushan2018deepjdot,balaji2019normalized,2018Deep,li2020generating}. The motivation is to collect unlabeled data from the target domain with no annotation cost on class label, and then adapt a model trained with the labeled source data to the unlabeled target data.

\begin{figure*}[t]
    \centering
    \subfloat[Static model]{\label{fig:visual_static}
        \includegraphics[trim=146 62 112 24,clip,width=0.5\textwidth]{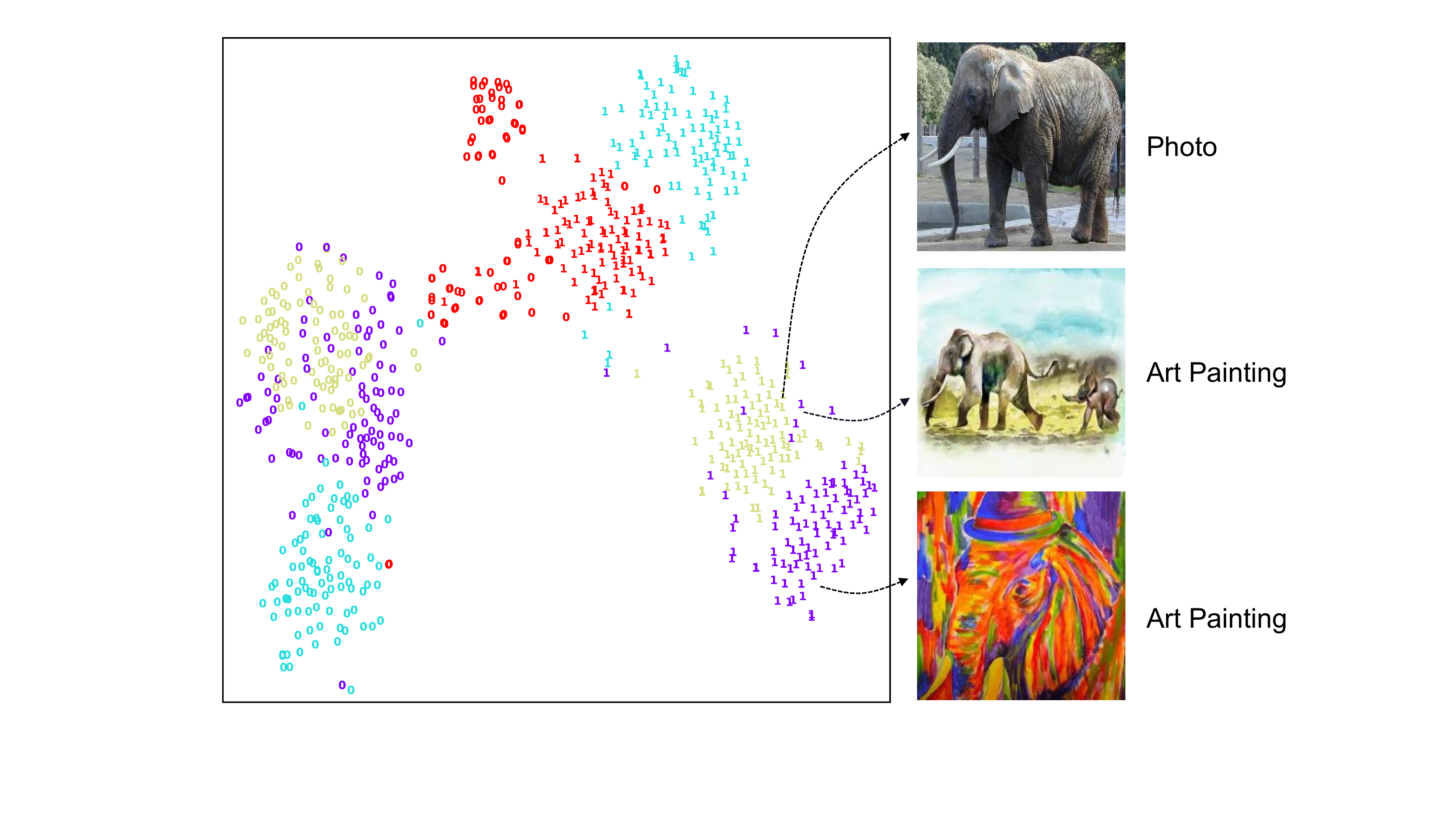}
    }
    ~
    \subfloat[DIDA-Net]{\label{fig:visual_dida}
        \includegraphics[trim=40 40 40 24,clip,width=0.35\textwidth]{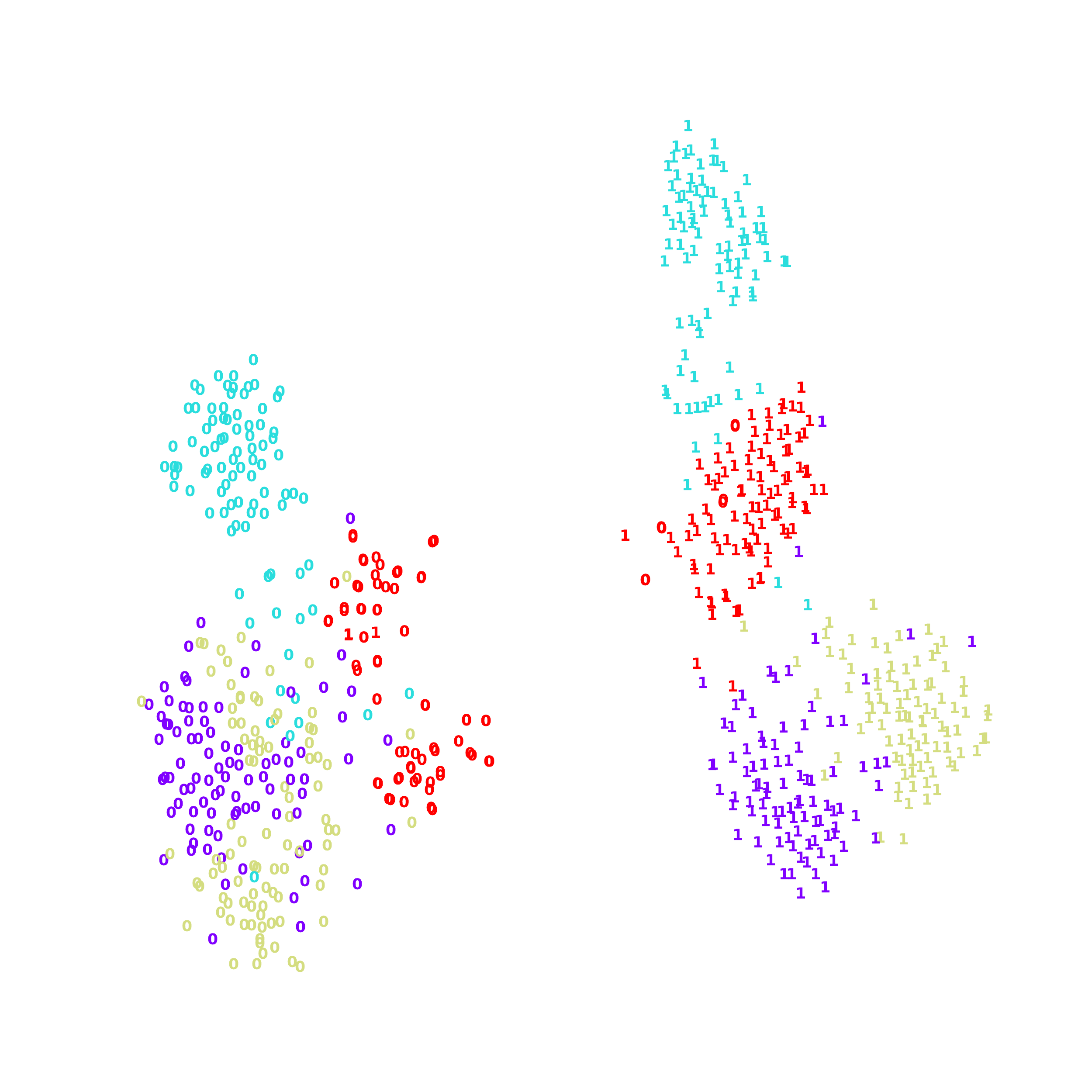}
    }
    \caption{
    2D t-SNE~\cite{maaten2008visualizing} visualization of features learned by (a) an existing feature alignment based MSDA method and (b) our DIDA-Net on PACS. Four colors denote four different domains. `0' and `1' denote the dog class and the elephant class respectively. (a) It is clear that some paintings are of drastically different styles, and a painting can be photo-realistic and distributed more closely to photos in the feature space.  (b) Our DIDA-Net can better align features of different domain in each class, leading to better class-wise separability.
    }
    \label{fig:intro_difc}
\end{figure*}

Early UDA works have focused on single-source domain adaptation~\cite{gretton2012kernel,tzeng2014deep,saito2017asymmetric}. Most single-source methods minimize the distribution discrepancy between source and target domains. This is usually achieved by using the domain labels for domain-specific feature alignment~\cite{gretton2012kernel,long2015learning,bhushan2018deepjdot,balaji2019normalized}. Nonetheless, it is more practical that source domain data are collected from multiple sources. Therefore, recent research has paid more attention to multi-source domain adaptation (MSDA)~\cite{Peng_2019_ICCV,wang2020learning,yang2020curriculum}. 
Most existing single-source and multi-source UDA methods assume that each domain data comes with both class and domain labels. 
The latter is used to perform domain-specific feature alignment based on the assumption that data samples collected from the same `domain’ follow the same distribution. Specifically, to learn a common feature space shared by source and target data, most methods align feature distributions across domains using a shared CNN model for feature extraction~\cite{Peng_2019_ICCV,zhao2018adversarial}. 

\begin{figure}
    \centering
    \includegraphics[trim=26 20 12 24,clip,width=0.7\columnwidth]{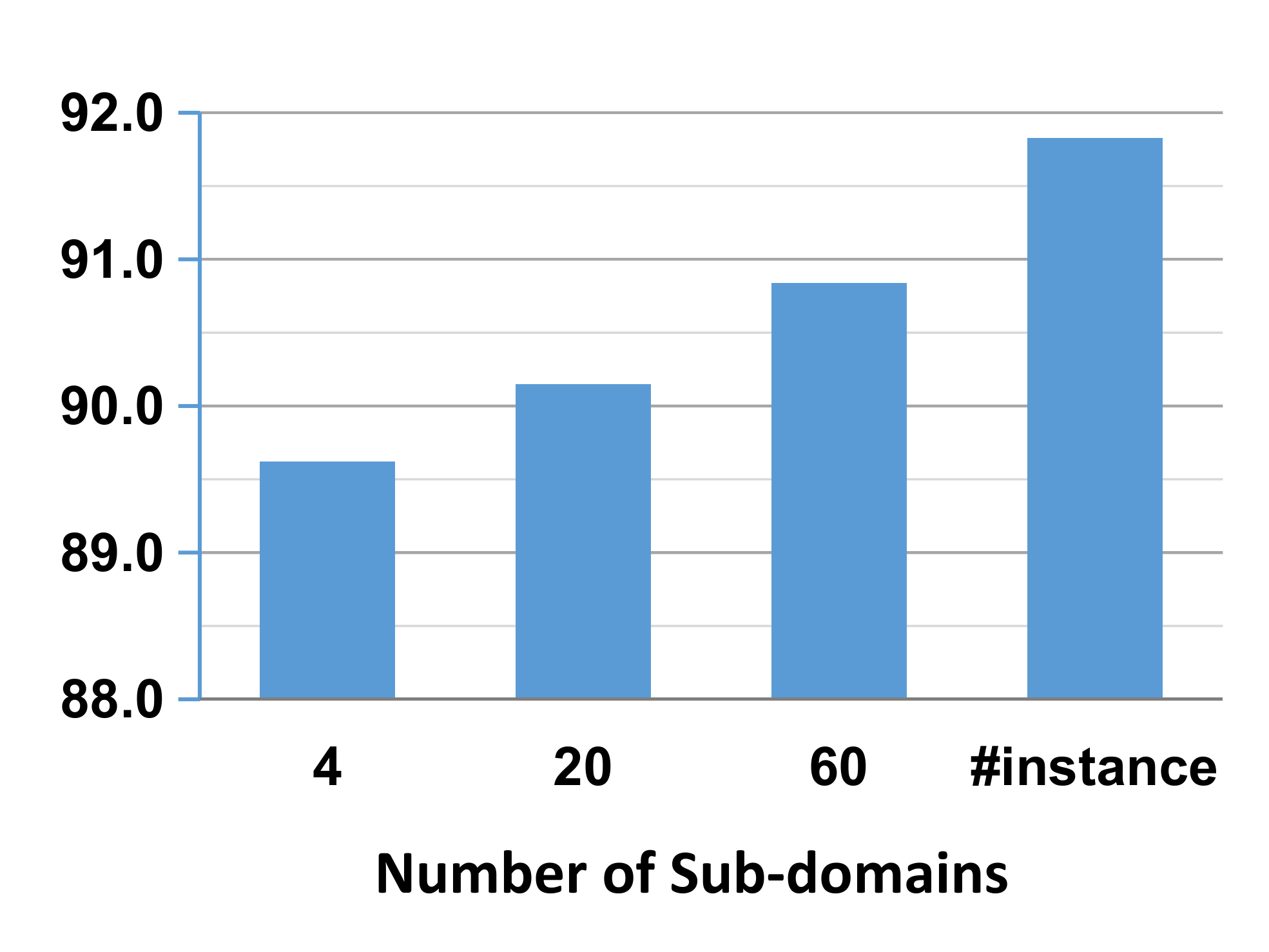}
    \caption{Accuracy on PACS with number of sub-domain increased. For \#Sub-domain=20 or 60, we adopt K-means to cluster features of different instances to obtain the sub-domain label for each instance.}
    \label{fig:acc_subdomain}
\end{figure}

\begin{figure}
    \centering
    \includegraphics[trim=35 7 2 0,clip,width=\columnwidth]{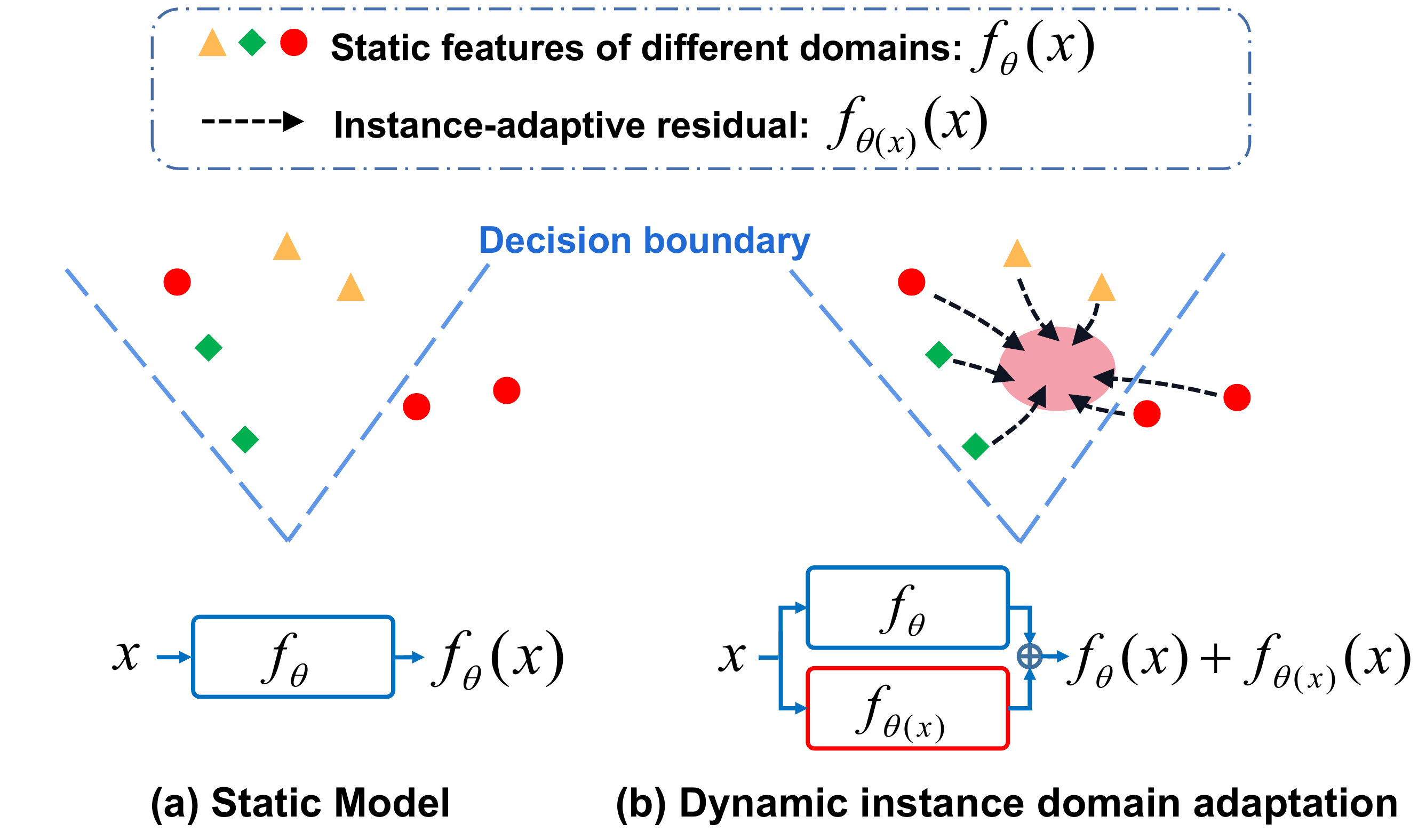}
    \caption{Static model vs. DIDA. (a) A static model $f_{\theta}$ aims to extract domain-agnostic features relying on domain annotations. However, when the target domain is too coarse-grained (e.g. the red dots) in single-source domain adaptation setting (or when there are multiple coarse-grained domains in MSDA), such a static model may fail to completely align all the domains. This can further result in some mis-classifications on the target domain, due to the domain shift and the absence of target labels. (b) Based on the static model $f_{\theta}$, our DIDA aims to better align all the domains in a specific class (the pink region) by introducing instance-adaptive residual. The residual $f_{\theta(x)}(x)$ is designed to be instance-adaptive because different instances usually have different styles, thus need different residuals for better class-wise alignment. So essentially the DIDA views each instance as a fine domain.}
    \label{fig:dida_residuals}
\end{figure}

However, this cross-domain feature distribution consistency assumption is typically incorrect, which in turn impedes the effectiveness of existing UDA methods. This is because the domains in the source/target data are often defined artificially and too coarse-grained. For example, in the popular PACS \cite{li2017deeper} benchmarking dataset, two of the four domains considered are photo and art painting. Fig.~\ref{fig:visual_static} shows that (1) this domain definition is too coarse; for example, dozens of modern painting styles have been developed, each differing dramatically from those of the classic styles. (2) The definition is too artificial and does not necessarily correspond to data distribution. For instance, some paintings are remarkably photo-realistic. Relying on the domain labels and enforcing distribution consistency across each domain can thus be counter-productive, as shown in Fig.~\ref{fig:dida_residuals}(a). One potential solution is more fine-grained domain annotation. 
As shown in Fig.~\ref{fig:acc_subdomain}, more fine-grained sub-domains lead to better domain adaptation results than coarse-grained domains (e.g. \#Sub-domain=4). When viewing each instance as a fine domain, we achieve the best result of 91.83\% on PACS. The best result owes to the better feature-level alignment of different domains in each class, which leads to a better class-wise separability, as depicted in Fig.~\ref{fig:visual_dida}.
However, domain definition is often subjective and requires expert knowledge (e.g., artistic training). 
In terms of annotation cost, one may be better off annotating the class labels of the target domain training data instead.

In this paper, we address both single-source and multi-source UDA from a completely new perspective, which is to \emph{view each instance as a fine domain}. With the granularity of domain pushed to the extreme and its definition becoming redundant, we cannot follow the existing feature alignment paradigm. Instead, we propose to perform \emph{dynamic instance domain adaptation} (DIDA). As shown in Fig.~\ref{fig:dida_residuals}(b), a dynamic neural network with multi-scale adaptive convolutional kernels is developed to generate \emph{instance-adaptive residuals} to adapt domain-agnostic deep features to each individual instance. This residual design reflects the nature of our problem: data samples from source and target are of the same labels but different in styles with the two parts modeled in two additive branches, i.e., $f_{\theta}$ and $f_{\theta(x)}$ in Fig.~\ref{fig:dida_residuals}(b).  Specifically, a DIDA module is attached to the top layer in a CNN as an auxiliary branch (suitable for any CNN architectures). It first uses a kernel generator to synthesize multi-scale convolutional kernels based on each specific instance, and then convolves the input features with these kernels to generate the instance-adaptive residuals.  Further, instead of imposing intricate feature alignment losses, we employ only a cross-entropy loss for both source and target data with pseudo labels utilized for the latter.

The contributions in this paper are summarized as follows. 
\begin{itemize}
    \item We propose a novel perspective to unsupervised domain adaptation, viewing each instance as a fine domain, and introduce an instance based domain adaptation.
    \item To realize instance based domain adaptation, we propose the DIDA module, which is based on dynamic neural networks for generating instance-adaptive residuals to calibrate deep CNN features.
    \item We conduct extensive experiments on two single-source domain adaptation datasets, including Digits and Office-Home, and three MSDA datasets, including DomainNet, Digit-Five, and PACS, and demonstrate the new state-of-the-art performance of our DIDA-Net on all the setups.
\end{itemize}

\section{Related Work}

\paragraph{Single-source domain adaptation}
Single-source domain adaptation aims to adapt a model from a labeled source domain to an unlabeled target domain. Most domain adaptation methods seek to minimize the distribution discrepancy between these two domains. The key lies in how to measure the distribution distance. Some methods use distance metrics, such as Maximum Mean Discrepancy (MMD)~\cite{gretton2012kernel,long2015learning,long2016unsupervised, tzeng2014deep}, optimal transport~\cite{bhushan2018deepjdot,balaji2019normalized}, and its variant ~\cite{li2020enhanced} with category prior integrated in, as well as the Kullback-Leibler (KL) divergence~\cite{zhuang2015supervised}, while others exploit adversarial training~\cite{2018Maximum, ganin2016domain}. All these methods learn a static model to produce domain-invariant features. This might be easy for the single source case but is more difficult for MSDA where source data distribution is more diverse.

\paragraph{Multi-source domain adaptation (MSDA)}
MSDA considers the scenarios where multiple source domains are available. Most MSDA methods are also based on distribution alignment using a static model. Deep cocktail network (DCTN)~\cite{2018Deep} learns a domain discriminator for each source-target pair. M$^3$SDA~\cite{Peng_2019_ICCV} applies moment matching to reducing the domain distance among source and target domains. LtC-MSDA~\cite{wang2020learning} explores the interactions of feature representations between different domains by constructing a knowledge graph to propagate the knowledge from source domains to the target. Curriculum manager for source selection (CMSS)~\cite{yang2020curriculum} selects source domains or samples that are easier to be aligned with the target based on their transferability. 
Pseudo Target Method for Multi-source Domain Adaptation (PTMDA)~\cite{ren2022multi} constructs pseudo target domain based on adversarial learning, and then uses pseudo target domains to align with source domains. 
Different from these methods that attempt to learn a common feature space using a static model, our DIDA-Net has dynamic model parameters conditioned on each instance for instance adaptation. No source domain labels are needed in our model. 

The most related work to our DIDA-Net is DRT~\cite{li2021dynamic}. DRT dynamically adapts the model for each sample by some injected residual blocks, while adopting the adversarial learning as per~\cite{2018Maximum} to align the source and target domains.
In contrast, our DIDA-Net views each instance as a fine domain, without using any artificially-defined domain label or domain alignment supervision. We also show that our DIDA-Net outperforms DRT in terms of both adaptation performance and compute efficiency as shown in Table~\ref{tab:compare_stateoftheart} and ~\ref{tab:dida_vs_drt}.

\begin{figure*}[t]
    \centering
    \includegraphics[trim=60 103 45 20,clip,width=\textwidth]{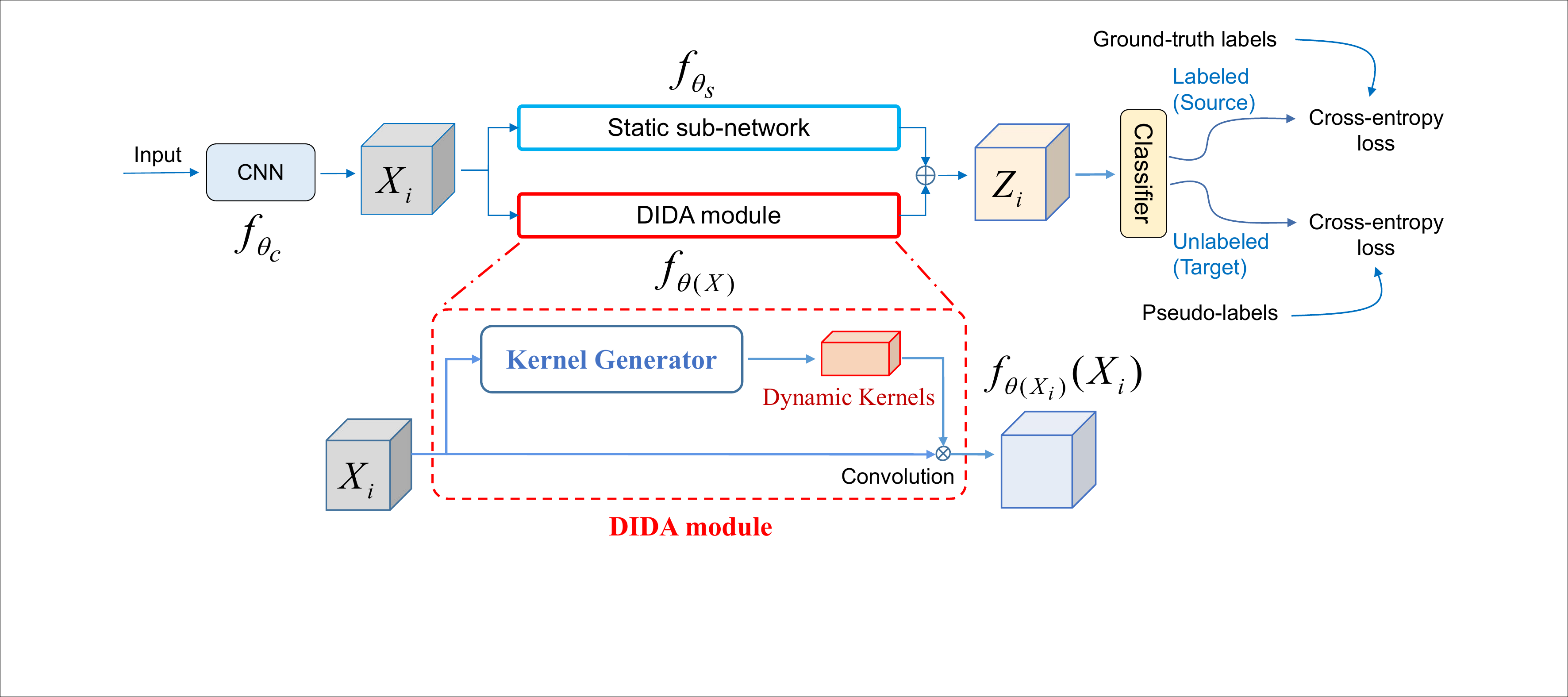}
    \caption{The architecture of our DIDA-Net. The final features of DIDA-Net, $Z_i$, are obtained by summing the outputs of a static sub-network $f_{\theta_s}$ and the DIDA module $f_{\theta(x)}$. The key component in the DIDA module is a kernel generator which generates dynamic convolutional kernels conditioned on the input.}
    \label{fig:difc_archi}
\end{figure*}

\paragraph{Dynamic neural networks}  
A common approach is to generate input-dependent weights for each convolutional kernel and sum up the weighted kernels as dynamic ensembled parameters~\cite{yang2019condconv,chen2020dynamic}. An alternative way is to dynamically generate convolutional kernels using a sub-network~\cite{jia2016dynamic,ha2016hypernetworks,he2019dynamic}. Instead of generating convolutional kernels, some methods re-scale features with instance-conditioned channel/spatial attention~\cite{hu2018squeeze,woo2018cbam,Wang_2017_CVPR,chen2017sca}. Other methods~\cite{wu2018unsupervised} use image features to formulate a non-parametric softmax classifier, which can be regarded as dynamic classifier, to distinguish each instance as a single class. In addition to dynamic parameters, dynamic computational graphs have also been explored, such as input-dependent layer dropout~\cite{wang2018skipnet,veit2018convolutional} or feature fusion~\cite{eigen2013learning,mullapudi2018hydranets}. Our DIDA-Net is essentially a dynamic neural network~\cite{jia2016dynamic} or hyper-network~\cite{ha2016hypernetworks}, which differs from existing models in using residual branch as an adapter. This design reflects the nature of our problem, i.e., data samples from source and target are of the same labels but different in styles with the two parts modeled in two additive branches.

\section{Methodology}

\subsection{Unsupervised Domain Adaptation}
The goal of unsupervised domain adaptation (UDA) is to adapt a model trained on $K$ source domains, $\{ \mathcal{S}_1, ..., \mathcal{S}_K \}$, to a target domain $\mathcal{T}$. For single-source setting, $K=1$; for MSDA, $K>1$. Each source domain contains labeled training instances $\mathcal{S}_k=\{(x_i^{\mathcal{S}_k}, y_i^{\mathcal{S}_k})\}_{i=1}^{N_{\mathcal{S}_k}}$, with $x$ and $y$ denoting image and label respectively. The target domain is assumed to share the same label space with the source domains, but all its training data are unlabeled, $\mathcal{T}=\{x_{i}^\mathcal{T}\}_{i=1}^{N_\mathcal{T}}$. These domains are manually defined, which means that the split of domains can be coarse. The focus of this paper is image classification problem, though the model can be used in other machine learning tasks. The objective is to improve the classification accuracy on the test set in the target domain.

\subsection{Dynamic Instance Domain Adaptation Network}
To better adapt a model from source domains to target one, we need to model both the domain styles and class information. The class information is used for the classification task while the style information is modeled to avoid bringing negative impact (or equivalently, to better align different domains in each class). The style information needs to be modeled because we cannot completely disentangle the class information from the style information, otherwise no domain adaptation is needed. Therefore, we aim to extract domain-agnostic class information and domain-specific style information, with the latter used for better class-wise alignment.

Since artificially-defined domain can be too coarse, we ignore the domain labels and view each instance as a fine domain instead. The instance-wise fine domain requires that the parameters of convolutional kernels should be conditioned on each instance (i.e., fine domain). This is because ~\cite{rozantsev2018beyond,bermudez2019domain} claim that different domains with different domain styles need different mappings for alignment in a common sub-space. These mappings are usually implemented as different convolution kernels, which are used for extracting different domain styles. These kernels are conditioned on different parameter groups, with each parameter group extracting a specific style pattern. Therefore, we essentially need different convolutional parameter groups for extracting the domain style of each instance if we view each instance as a fine domain. In other words, the parameters of kernels should be conditioned on each instance.

Our objective is to first extract a domain-agnostic feature representation and then adapt it to each individual data sample.  
To that end, we introduce a dynamic instance domain adaptation network (DIDA-Net). The DIDA-Net outputs instance-conditioned features with dynamically generated convolutional kernels. With these kernels, class-wise feature compactness across source and target data can be implicitly achieved by using a shared classifier across domains. 

The architecture of our DIDA-Net is depicted in Fig.~\ref{fig:difc_archi}. We first sample $K\times B$ images from the source domains and $B$ images from the target respectively, i.e., $\{x_i^{\mathcal{S}}\}_{i=1}^{KB},\{x_i^{\mathcal{T}}\}_{i=1}^B$ where $\mathcal{S}=\mathcal{S}_1\cup ...\cup \mathcal{S}_K$. These images are input to the same feature extractor and classifier. The feature extractor consists of two different static sub-networks $f_{\theta_c}$ and $f_{\theta_s}$, and a dynamic instance domain adaptation module (DIDA module) $f_{\theta(X)}$. The static CNN sub-network $f_{\theta_c}$ is used to extract intermediate features $X_i\in \mathbb{R}^{h \times w \times c}$ for $x_i$ (domain-specific superscript omitted for brevity), where $h,w,c$ denote width, height and channel depth respectively. The feature map $X_i$ is then fed forward to a static sub-network $f_{\theta_s}$ and a DIDA module $f_{\theta(x)}$, with the static sub-network modeling instance-shared class information and the DIDA module modeling instance-dependent styles. The DIDA module has instance-dependent convolutional parameters $\theta(x)$, which are generated by a kernel generator, for instance adaptation.
Finally, we sum the outputs of $f_{\theta_s}$ and $f_{\theta(X)}$ to obtain final feature representations $Z_i$:
\begin{equation} \label{eq:difc}
    Z_i=f_{\theta_s}(X_i) + f_{\theta(X_i)}(X_i),
\end{equation}
where $f_{\theta(X_i)}(X_i)$ is dynamic residual compensations for each instance. Due to the dynamic residual compensations from the DIDA module, $Z_i$ is instance-adaptive. Below we detail the design of the DIDA module.


\begin{figure*}[t]
    \centering
    \includegraphics[trim=25 23 20 21,clip,width=1\textwidth]{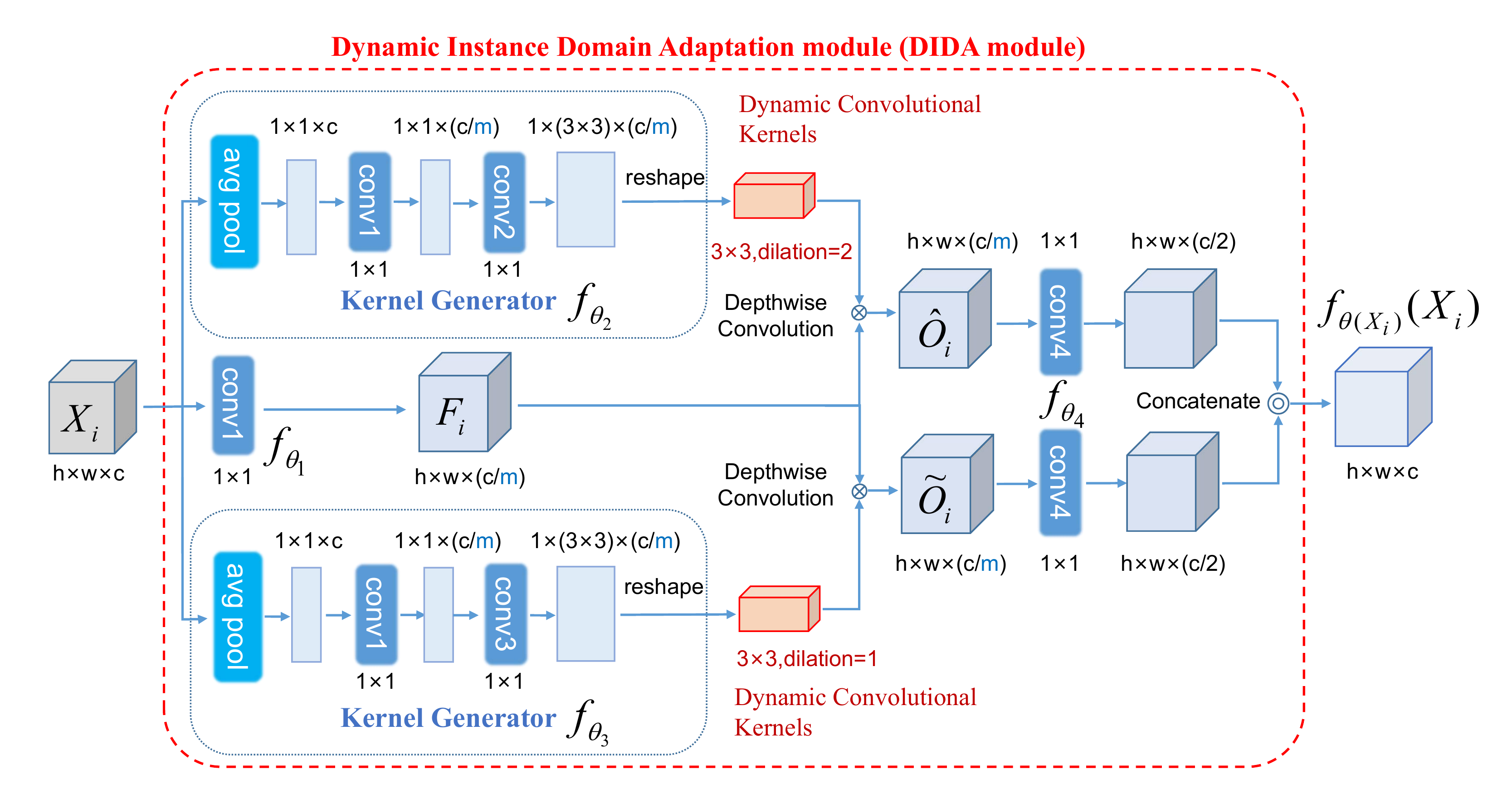}
    \caption{A schematic of the dynamic instance domain adaptation module (DIDA module). 
     The DIDA module has three branches for the input. The middle branch reduces channels of the input. The top and bottom branches are kernel generator branches. These kernel generators generate convolutional kernels of different dilation rates to capture objects of different scales. Then, the generated dynamic kernels convolve with the channel-reduced input in a depth-wise way to output $\hat{O}_i,\tilde{O}_i$. After that, the channels of $\hat{O}_i,\tilde{O}_i$ are increased and concatenated. The concatenated output is the dynamic residual compensations for the static features of corresponding input instance. The DIDA module is lightweight, with only four $1\times 1$ convolution layers, i.e., `conv1'$\sim$`conv4'.}
    \label{fig:idam_archi}
\end{figure*}

\paragraph{Dynamic instance domain adaptation module} We follow the following principle when designing the DIDA module: instance-conditioned module that can tackle the scale variation of objects in an efficient way. As such, generating dynamic kernels of different dilation rates is more suitable for the object's scale issue than re-scaling features (e.g. attention module~\cite{hu2018squeeze,woo2018cbam,zuo2021attention}) or weighted sum of several kernels~\cite{li2021dynamic}, and usually more efficient than dynamic computational graphs (e.g., feature fusion may need multiple backbones for different domains~\cite{rozantsev2018beyond,bermudez2019domain}).

As shown in Fig.~\ref{fig:idam_archi}, the DIDA module has three branches for the input. The middle branch reduces the channels of the input by $m$, where $m$ is channel reduction rate. The channel reduction is operated by a $1\times1$ convolution layer $f_{\theta_1}$ (i.e., `conv1' in Fig.~\ref{fig:idam_archi}), with $\theta_1$ as its parameters.
The output of the middle branch is $F_i=f_{\theta_1}(X_i)\in \mathbb{R}^{h \times w \times (c/m)}$. 

The top and bottom branches are kernel generator branches. 
The kernel generators $f_{\theta_2}, f_{\theta_3}$ generate $3\times 3$ convolutional kernels of different dilation rates. The generated kernels are denoted as $f_{\theta_2}(X_i)$ and $f_{\theta_3}(X_i)$ respectively. We use different dilation rates to capture objects of different scales, since objects in different domains usually have different scales (see Fig.~\ref{fig:sample_imgs}(a)). Then, we have
\begin{align}
    \hat{O}_i&=f_{\theta_2}(X_i)\otimes F_i=f_{\theta_2}(X_i)\otimes f_{\theta_1}(X_i),\\
    \tilde{O}_i&=f_{\theta_3}(X_i)\otimes F_i=f_{\theta_3}(X_i)\otimes f_{\theta_1}(X_i),
\end{align}
where symbol $\otimes$ refers to depth-wise convolution, and $\hat{O}_i,\tilde{O}_i$ are the output features. Note that the dynamic convolutional kernels $f_{\theta_2}(X_i),f_{\theta_3}(X_i)$ are dependent on the input instance $X_i$, so they contribute to instance adaptation.

In our implementation, a kernel generator comprises of a global average pooling layer, a channel reduction layer and a kernel generation layer. The average pooling layer integrates spatial context by squeezing the spatial dimension. The channel reduction layer is shared among three branches, i.e., $f_{\theta_1}$, to avoid computational overload. The kernel generation layer, i.e., `conv2'/`conv3' in Fig.~\ref{fig:idam_archi}, increases the elements of the feature map \footnote{This operation needs dimension swap for normal convolution. We first swap the Width (W) with Channel (C) of the input: 1$\times$(c/m)$\times$1, then use conv2/conv3 to increase its channel to 1$\times$(c/m)$\times$(3$\times$3). Finally, we get 1$\times$(3$\times$3)$\times$(c/m) by swapping the W, C dimension back.} and its output is reshaped as the generated kernels.

Finally, we use a convolutional layer $f_{\theta_4}$ (`conv4' in Fig.~\ref{fig:idam_archi}) to increase the channel of $\hat{O}_i,\tilde{O}_i$, and concatenate the outputs as the dynamic residuals:
\begin{align} 
    f_{\theta(X_i)}(X_i)&=[f_{\theta_4}(\hat{O}_i);f_{\theta_4}(\tilde{O}_i)]\\
    &= [f_{\theta_4}(f_{\theta_2}(X_i)\otimes f_{\theta_1}(X_i)); \\
    &\quad f_{\theta_4}(f_{\theta_3}(X_i)\otimes f_{\theta_1}(X_i))],
\end{align}
where $[f_{\theta_4}(\hat{O}_i);f_{\theta_4}(\tilde{O}_i)]$ denotes channel-wise concatenation of the two inputs. Note that the DIDA module $f_{\theta(X_i)}$ is input-dependent due to its instance-conditioned convolutional kernels $f_{\theta_2}(X_i),f_{\theta_3}(X_i)$.

\subsection{Learning}

We leverage a shared classifier across domains to enforce the final features $Z_i$ to be both domain-invariant and discriminative. The classifier is supervised by cross-entropy loss for both labeled source and pseudo-labeled target domain training data.

\paragraph{Supervised learning for labeled source data}
We apply cross-entropy loss to labeled source data, i.e.,
\begin{equation} \label{eq:Ls}
L_s = -\frac{1}{KB} \sum_{i=1}^{KB} \log p_{i, y_i^{\mathcal{S}}}^{\mathcal{S}},
\end{equation}
where $p_{i, y_i}^{\mathcal{S}}$ is the $y_i^{\mathcal{S}}$-th (ground-truth) prediction of $p_i^{\mathcal{S}}$. $p_i^{\mathcal{S}}$ denotes the output probability of the shared classifier for $x_i^{\mathcal{S}}$.

\paragraph{Pseudo-labeling for unlabeled target data} Pseudo-labeling has shown its effectiveness of improving domain adaptation performance~\cite{shin2020two, gu2020spherical}. Instead of designing a new pseudo-labeling technique, we reuse the method proposed for semi-supervised learning in FixMatch~\cite{sohn2020fixmatch} for unlabeled target data. We feed a weakly-augmented version of a target image to the network and obtain its predicted class $\hat{y}_i^{\mathcal{T}}$. The prediction is accepted as pseudo label only if it has high confidence, e.g., $\hat{y}_i^{\mathcal{T}}>\tau$. The threshold $\tau$ is 0.95 in this paper. Then, we impose the cross-entropy loss on the classifier's output of a strongly-augmented version of the same image:
\begin{equation} \label{eq:Lt}
L_t = -\frac{1}{B} \sum_{i=1}^B \mathbbm{1}(q(\hat{y}_i^{\mathcal{T}}) \geq \tau) \log p_{i, \hat{y}_i^{\mathcal{T}}}^{\mathcal{T}},
\end{equation}
where $\mathbbm{1}(\cdot)$ is the indicator function, and $q(\hat{y}_i^{\mathcal{T}})$ is the predicted probability on the $\hat{y}_i^{\mathcal{T}}$-th class.

\paragraph{Training}
We impose the losses in Eqs.~(\ref{eq:Ls}) and~(\ref{eq:Lt}) on the \emph{shared classifier} for end-to-end training, i.e.,
\begin{equation} \label{eq:final_loss}
L = L_s +  L_t,
\end{equation}
where $L$ is the overall loss. Minimizing $L$ can lead to the source and target features implicitly aligned in each class, which we call it as implicit class-wise domain alignment.

\textbf{Discussion on implicit class-wise domain alignment.} Minimizing Eq.~\eqref{eq:final_loss} enforces final features $Z_i^{\mathcal{S}}$ and $Z_i^{\mathcal{T}}$ to lie in the same common latent space to fit the \emph{shared} classifier. More importantly, the features of the source and target domain, $Z_i^{\mathcal{S}}$ and $Z_i^{\mathcal{T}}$, are forced to be separable by using the same set of decision boundaries, due to the shared classifier. Accordingly, $Z_i^{\mathcal{S}}$ and $Z_i^{\mathcal{T}}$ are implicitly aligned in each class, leading to class-wise domain-invariant feature learning. This is much easier to achieve than previous explicit domain alignment loss since we emphasize the class-wise separability instead of completely aligning different domains. Therefore, we can alleviate the counter-productive classification performance on a single domain, even though different domains are not completely aligned. 
The effectiveness of such an implicit alignment is also validated by the visualization in Fig.~\ref{fig:visual_difc}.

\section{Experiments}

\subsection{Experimental Setting}
\label{sec:experiment_setting}

\begin{figure*}[t]
    \centering
    \includegraphics[trim=5 10 15 20,clip,width=\textwidth]{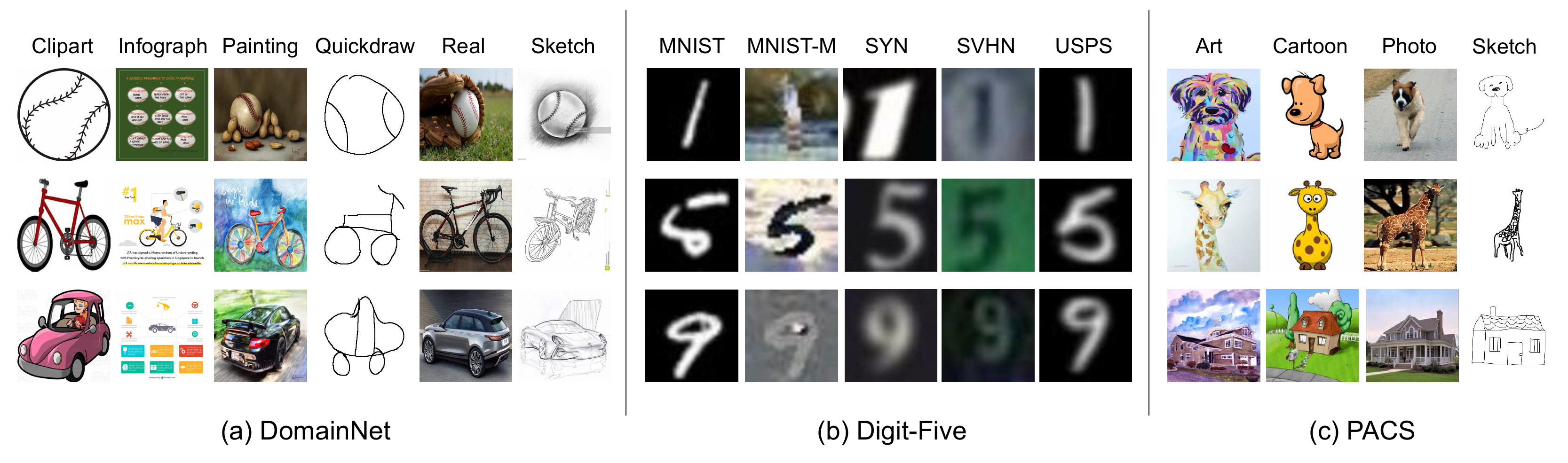}
    \caption{
    Example images from three MSDA datasets manifesting different types of domain shift.
    }
    \label{fig:sample_imgs}
\end{figure*}

\paragraph{Single-source domain adaptation}
\textbf{Datasets and protocols.} We adopt Digits and Office-Home for evaluation. 
Digits dataset contains images of 10 digit numbers. It has three domain transfer tasks: SVHN~\cite{37648} $\rightarrow$ MNIST~\cite{1998Gradient}, USPS $\rightarrow$ MNIST, MNIST $\rightarrow$ USPS.
Office-Home~\cite{venkateswara2017deep} includes 15,500 images from 65 object categories. These images are divided into four domains: Artistic (Ar), Clip Art (Cl), Product (Pr) and Real-World (Rw). It is worth noting that the artificially-defined domain in Office-Home can be coarse-grained. For example, the Artistic domain contains sketches, paintings and/or artistic depictions. We evaluate the DIDA-Net on all the 12 transfer tasks as in~\cite{tang2020unsupervised}.

\textbf{Training details.} On Digits, following ~\cite{2018Maximum}, we use a CNN backbone with three convolution layers and two fully connected layers for the SVHN $\rightarrow$ MNIST task, and a CNN with two convolution layers + two fully connected layers for USPS $\rightarrow$ MNIST, and MNIST $\rightarrow$ USPS tasks. For training, we use Adam~\cite{kingma2014adam} optimizer with learning rate of 2e-4, mini-batch size of 128.
On Office-Home, ImageNet-pretrained ResNet-50 is used as backbone, which is the same as ~\cite{tang2020unsupervised}. The model is then optimized using momentum-based SGD for 100 epochs, with an initial learning rate of 0.001 and mini-batch of 32.

\textbf{Model details.} For the CNN models used on Digits, we add the output of DIDA module to that of the last convolutional layer. For the ResNet-50 used on Office-Home, we insert the DIDA module after the third residual block and sum its output with the output of the fourth residual block (i.e. the static sub-network $f_{\theta_s}$ is the fourth residual block). Furthermore, to alleviate over-fitting of the DIDA-Net on single-source data, we apply the MixStyle~\cite{zhou2021domain} after the first convolutional layer (for Digits) or the first residual block (for Office-Home).

\paragraph{Multi-source domain adaptation (MSDA)}
\textbf{Datasets and protocols.} We evaluate our DIDA-Net on DomainNet, Digit-Five and PACS. 
DomainNet~\cite{Peng_2019_ICCV} is the largest, and most challenging, MSDA dataset so far with around 0.6 million images of 345 categories. It consists of six domains: Clipart, Infograph, Painting, Quickdraw, Real, and Sketch. The domain shift is dramatic due to large variation in object scale, image style, background, etc. 
Digit-Five comprises five digit datasets (MNIST~\cite{1998Gradient}, MNIST-M~\cite{syn_digits}, Synthetic Digits~\cite{syn_digits}, SVHN~\cite{37648}, and USPS). The domain shift lies in different font styles and backgrounds. Following~\cite{Peng_2019_ICCV}, we use all 9,298 images in USPS as a source domain for training; on the other four datasets, we sample 25,000 images from each domain for training and 9,000 images for testing. 
PACS~\cite{li2017deeper} has four distinct domains---Art Painting, Cartoon, Photo, and Sketch---with domain shift mainly corresponding to image style changes. There are 9,991 images of 7 classes. The train-val slits provided by~\cite{li2017deeper} are used. 
Sample images from these three datasets are shown in Fig.~\ref{fig:sample_imgs}.

\textbf{Training details.} On DomainNet and Digit-Five, we use the SGD with momentum as the optimizer and the cosine annealing rule~\cite{loshchilov2016sgdr} for decaying the learning rate; On PACS, we use the Adam optimizer~\cite{kingma2014adam}.
On DomainNet, we follow~\cite{Peng_2019_ICCV} and use the ImageNet-pretrained ResNet-101~\cite{he2016deep} as the CNN backbone. The model is trained for 40 epochs with the initial learning rate of 0.002. We set $B=6$ for each minibatch.
On Digit-Five, we adopt the same backbone as in~\cite{Peng_2019_ICCV} (three convolution layers + two fully connected layers). The batch size $B$ is set to 64. The model is trained for 30 epochs with the initial learning rate of 0.05.
On PACS, we use ResNet-18~\cite{he2016deep} as the CNN backbone, following~\cite{wang2020learning}. We set $B$ to 16, the initial learning rate to 5e-4, and the total training epoch to 100.

\textbf{Model details.} For ResNet used on PACS~\cite{li2017deeper} and DomainNet~\cite{Peng_2019_ICCV}, the static sub-network $f_{\theta_s}$ refers to the last residual block (denoted as \texttt{res4}) while $f_{\theta_c}$ represents the remaining three residual blocks (\texttt{res1-3}). The DIDA module is inserted after the third residual block. In addition, the learning rate for the DIDA module is set to $10 \times$ the global learning rate as it is trained from scratch. For the five-layer CNN model used on Digit-Five, the third convolution layer is seen as the static sub-network.

The channel reduction rate $m$ is set to 16 as default for both single-source and multi-source domain adaptation. Our experiments are conducted using PyTorch~\cite{paszke2017pytorch,paszke2019pytorch} and Dassl.pytorch~\cite{zhou2021dael} \footnote{\url{https://github.com/KaiyangZhou/Dassl.pytorch}}. The code is available at \url{https://github.com/Zhongying-Deng/DIDA}.

\subsection{Comparisons with the State of the Art}
~\label{subsec:compare_sota}

\paragraph{Single-source domain adaptation}

We first evaluate our DIDA-Net under single-source domain adaptation setting. The results on Digits and Office-Home are shown in Table~\ref{tab:single_src_digits} and Table~\ref{tab:single_src_office_home} respectively. We can see that our DIDA-Net obtains the state-of-the-art average accuracy on both datasets.

\textbf{Digits} In Table~\ref{tab:single_src_digits}, all the other competitors rely on the coarse-grained domain labels to force a static model to output domain-agnostic features. Among them, DANN~\cite{ganin2016domain}, MCD~\cite{2018Maximum}, LWC~\cite{ye2020light}, and ILA-DA+CDAN~\cite{sharma2021instance} adopt adversarial training for the domain-agnostic features while CKB~\cite{luo2021conditional} utilizes distance metrics. Different from these methods, our DIDA-Net treats each instance as a fine domain and exploits dynamic model for instance adaptation. In terms of the performance, the DIDA-Net achieves 2\% better than DANN, MCD, and ILA-DA+CDAN and comparable results with LWC, which demonstrates the efficacy of our DIDA-Net. 

\textbf{Office-Home} Table~\ref{tab:single_src_office_home} shows that our DIDA-Net outperforms the other domain alignment methods, such as SymNets~\cite{zhang2019domain}, RSDA-MSTN~\cite{gu2020spherical} and SRDC~\cite{tang2020unsupervised}.
Notably, the first three columns in Table~\ref{tab:single_src_office_home} show the results of other domains transfer to the coarse-grained Artistic domain (containing sketches, paintings and/or artistic depictions). When the coarse-grained Artistic is the target domain, the DIDA-Net surpasses all the other methods, with up to 2.2\% performance gap. This supports our claims: 1) feature distribution alignment across each coarse-grained domain can be suboptimal; 2) our DIDA-Net can effectively alleviate the coarse-grained domain issue by viewing each instance as a fine domain.

\begin{table}[t]
    \centering
    \caption{Single-source domain adaptation results on Digits.}
    \label{tab:single_src_digits}
    \begin{tabular}{l|ccc|c}
    \toprule
      \textbf{Method} & \textbf{M$\rightarrow$ U}  &\textbf{U$\rightarrow$ M} &\textbf{S$\rightarrow$ M} & \textbf{Avg}\\
    \midrule
       DANN~\cite{ganin2016domain} &90.8 &94.0 &83.1 &89.3\\
       MCD~\cite{2018Maximum} &94.2 &94.1 &96.2 &94.8\\
       LWC~\cite{ye2020light} &95.6 &97.1 &\textbf{97.1} &96.6\\
       CKB~\cite{luo2021conditional} &96.3 &96.6 &- &-\\
       ILA-DA+CDAN~\cite{sharma2021instance} &94.9 &97.5 &92.3 &94.9\\
    \midrule
       DIDA-Net (\emph{ours}) &\textbf{98.0}	&\textbf{97.9}	&94.8	&\textbf{96.9}\\
    \bottomrule
    \end{tabular}
\end{table}

\begin{table*}[t]
    \centering
    \caption{Single-source domain adaptation results on Office-Home.}
    \label{tab:single_src_office_home}
    \resizebox{\textwidth}{!}{
    \begin{tabular}{l|cccccccccccc|c}
    \toprule
      \textbf{Method} & Pr$\rightarrow$Ar  &Rw$\rightarrow$Ar &Cl$\rightarrow$Ar & Ar$\rightarrow$Cl & Rw$\rightarrow$Cl & Pr$\rightarrow$Cl & Ar$\rightarrow$Pr  & Cl$\rightarrow$Pr & Rw$\rightarrow$Pr & Ar$\rightarrow$Rw & Cl$\rightarrow$Rw  & Pr$\rightarrow$Rw &\textbf{Avg}\\
    \midrule
       ResNet-50~\cite{he2016deep} &38.5	&53.9	&37.4	&34.9	&41.2	&31.2	&50.0	&41.9	&59.9	&58.0	&46.2	&60.4	&46.1\\
       DANN~\cite{ganin2016domain} &46.1	&63.2	&47.0	&45.6	&51.8	&43.7	&59.3	&58.5	&76.8	&70.1	&60.9	&68.5	&57.6\\
       CDAN~\cite{long2017conditional} &55.6	&68.4	&54.4	&49.0	&55.4	&48.3	&69.3	&66.0	&80.5	&74.5	&68.4	&75.9	&63.8\\
       SymNets~\cite{zhang2019domain} &63.6	&73.8	&64.2	&47.7	&50.8	&47.6	&72.9	&71.3	&82.6	&78.5	&74.2	&79.4	&67.2\\
       RSDA-MSTN~\cite{gu2020spherical} &67.9	&75.8	&66.4	&53.2	&57.8	&53.0	&\textbf{77.7}	&74.0	&\textbf{85.4}	&\textbf{81.3}	&76.5	&\textbf{82.0}	&70.9\\
       SRDC~\cite{tang2020unsupervised} &68.7	&76.3	&69.5	&52.3	&57.1	&53.8	&76.3	&\textbf{76.2}	&85.0	&81.0	&\textbf{78.0}	&81.7	&71.3\\
       \midrule
       DIDA-Net (\emph{ours}) &\textbf{70.9}	&\textbf{78.5}	&\textbf{69.8}	&\textbf{59.4}	&\textbf{62.1}	&\textbf{58.6}	&76.0	&74.9	&84.5	&78.9	&73.6	&78.7 &\textbf{72.1} \\
    \bottomrule
    \end{tabular}
    }
\end{table*}

\begin{table*}[t]
    \caption{Results on three MSDA datasets. Our DIDA-Net outperforms the other state-of-the-art methods on all the datasets.}
    \vspace{-.4cm}
    \label{tab:compare_stateoftheart}
    \centering
\subfloat[Multi-source domain adaptation result on PACS. * denotes our implementation.]{\label{tab:pacs}
    \tabstyle{13pt}
    \begin{tabular}{l|cccc|c}
    \toprule
    \textbf{Method} & \textbf{Art Painting} & \textbf{Cartoon} & \textbf{Sketch} & \textbf{Photo} & \textbf{Avg}\\
    \midrule 
    Source-only &81.22	&78.54	&72.54	&95.45	&81.94\\
    MDAN~\cite{zhao2018adversarial} &83.54 &82.34 &72.42 &92.91 &82.80\\
    DCTN~\cite{2018Deep}  &84.67 &86.72 &71.84 &95.60 &84.71\\
    M$^3$SDA-$\beta$~\cite{Peng_2019_ICCV}  &84.20 &85.68 &74.62 &94.47 &84.74\\
    MDDA~\cite{zhao2020multi} &86.73 &86.24 &77.56 &93.89 &86.11\\
    LtC-MSDA~\cite{wang2020learning} &90.19 &90.47 &81.53 &97.23 &89.85\\
    DRT\textsuperscript{*}~\cite{li2021dynamic} &88.26 &89.71 &74.86 &96.38 &87.30\\
    \midrule
    {DIDA-Net} (\emph{ours}) &\textbf{93.39}	&\textbf{90.81}	&\textbf{84.77}	&\textbf{98.36}	&\textbf{91.83} \\
    \bottomrule
    \end{tabular}
}
\\
\subfloat[Multi-source domain adaptation result on Digit-Five.]{
    \begin{tabular}{l|ccccc|c}
    \toprule
    \textbf{Method} & \textbf{MNIST} & \textbf{USPS} & \textbf{MNIST-M} & \textbf{SVHN} & \textbf{Synthetic} & \textbf{Avg}\\
    \midrule 
    Source-only~\cite{yang2020curriculum}  & 92.3$\pm$0.91 &90.7$\pm$0.54  &63.7$\pm$0.83 &71.5$\pm$0.75 &83.4$\pm$0.79 &80.3\\ 
    DANN~\cite{ganin2016domain} &97.9$\pm$0.83 &93.4$\pm$0.79 &70.8$\pm$0.94 &68.5$\pm$0.85 &87.3$\pm$0.68 &83.6\\ 
    DCTN~\cite{2018Deep} & 96.2$\pm$0.80  &92.8$\pm$0.30 &70.5$\pm$1.20 &77.6$\pm$0.40 &86.8$\pm$0.80 &84.8 \\
    MCD~\cite{2018Maximum}  & 96.2$\pm$0.81 &95.3$\pm$0.74 &72.5$\pm$0.67 &78.8$\pm$0.78 &87.4$\pm$0.65 &86.1\\ 
    M$^3$SDA-$\beta$~\cite{Peng_2019_ICCV}   &  98.4$\pm$0.68 &96.1$\pm$0.81 &72.8$\pm$1.13 &81.3$\pm$0.86 &89.6$\pm$0.56 &87.6\\ 
    CMSS~\cite{yang2020curriculum} &99.0$\pm$0.08 &97.7$\pm$0.13 &75.3$\pm$0.57 &88.4$\pm$0.54 &93.7$\pm$0.21 &90.8 \\ 
    LtC-MSDA~\cite{wang2020learning}   &99.0$\pm$0.40 &98.3$\pm$0.40 &85.6$\pm$0.80 &83.2$\pm$0.60 &93.0$\pm$0.50 &91.8\\
    DRT~\cite{li2021dynamic} &\textbf{99.3}$\pm$0.05 &98.4$\pm$0.12 &81.0$\pm$0.34 &86.7$\pm$0.38 &93.9$\pm$0.34 &91.9\\
    \midrule
    {DIDA-Net} (\emph{ours}) &\textbf{99.3}$\pm$0.07 &\textbf{98.6}$\pm$0.10  &\textbf{85.7}$\pm$0.12  &\textbf{91.7}$\pm$0.08  &\textbf{97.3}$\pm$0.01 & \textbf{94.5} \\ 
    \bottomrule
    \end{tabular}
    \label{tab:digit-five}
}
\\
\subfloat[Multi-source domain adaptation result on DomainNet.]{
    \begin{tabular}{l|cccccc|c}
    \toprule
        \textbf{Method} & \textbf{Clipart} & \textbf{Infograph} & \textbf{Painting} & \textbf{Quickdraw} & \textbf{Real}  & \textbf{Sketch} & \textbf{Avg}\\
    \midrule 
        Source-only~\cite{Peng_2019_ICCV} &47.6$\pm$0.52 &13.0$\pm$0.41 & 38.1$\pm$0.45 &13.3$\pm$0.39 &51.9$\pm$0.85 &33.7$\pm$0.54 &32.9\\
        DANN~\cite{ganin2016domain} &45.5$\pm$0.59 &13.1$\pm$0.72 &37.0$\pm$0.69 &13.2$\pm$0.77 &48.9$\pm$0.65 &31.8$\pm$0.62 &32.6\\
        DCTN~\cite{2018Deep} &48.6$\pm$0.73 &23.5$\pm$0.59 &48.8$\pm$0.63 &7.2$\pm$0.46 &53.5$\pm$0.56 &47.3$\pm$0.47 &38.2 \\
        MCD~\cite{2018Maximum} &54.3$\pm$0.64 &22.1$\pm$0.70 &45.7$\pm$0.63 &7.6$\pm$0.49 &58.4$\pm$0.65 &43.5$\pm$0.57 &38.5\\
        M$^3$SDA-$\beta$~\cite{Peng_2019_ICCV} &58.6$\pm$0.53 &26.0$\pm$0.89 &52.3$\pm$0.55 &6.3$\pm$0.58 &62.7$\pm$0.51 &49.5$\pm$0.76 &42.6\\
        CMSS~\cite{yang2020curriculum} &64.2$\pm$0.18 &28.0$\pm$0.20 &53.6$\pm$0.39 &16.0$\pm$0.12 &63.4$\pm$0.21 &53.8$\pm$0.35 &46.5\\
        LtC-MSDA~\cite{wang2020learning} &63.1$\pm$0.50 &28.7$\pm$0.70 &56.1$\pm$0.50 &16.3$\pm$0.50 &66.1$\pm$0.60 &53.8$\pm$0.60 &47.4\\
        DRT~\cite{li2021dynamic} &69.7$\pm$0.24 &31.0$\pm$0.56 &59.5$\pm$0.43 &9.9$\pm$1.03 &68.4$\pm$0.28 &59.4$\pm$0.21 &49.7\\
        DRT+Self-Training &71.0$\pm$0.21 &\textbf{31.6}$\pm$0.44 &\textbf{61.0}$\pm$0.32 &12.3$\pm$0.38 &\textbf{71.4}$\pm$0.23 &\textbf{60.7}$\pm$0.31 &51.3\\
    \midrule
        DIDA-Net (\emph{ours}) &\textbf{73.6}$\pm$0.10 &28.6$\pm$0.25 &58.7$\pm$0.13 &\textbf{21.1}$\pm$0.36 &68.9$\pm$0.27 &60.4$\pm$0.14 &\textbf{51.9}\\
    \bottomrule
    \end{tabular}
    \label{tab:domainnet}
}
\end{table*}

\paragraph{Multi-source domain adaptation}
Table~\ref{tab:compare_stateoftheart} compares our DIDA-Net with the state of the art on the three MSDA datasets. Note that we combine all the source domains as a single one when applying single-source domain adaptation methods, such as DANN~\cite{ganin2016domain} and MCD~\cite{2018Maximum}, to multi-source scenario. This is consistent with what previous multi-source domain adaptation works~\cite{Peng_2019_ICCV,wang2020learning} did. It is clear that DIDA-Net outperforms the other state-of-the-art methods on all datasets. Next we give some more detailed discussions.

\textbf{Digit-Five and PACS}
Except DIDA-Net, all the other MSDA methods are based on either domain-level alignment or a static model. MCD~\cite{2018Maximum} and M$^3$SDA-$\beta$~\cite{Peng_2019_ICCV} are top-performing competitors, which exploit task-specific classifiers for fine alignment. Our DIDA-Net also investigates the idea of fine alignment, but targets the instance level, which better suits real-world MSDA datasets with complicated data distributions. The gap obtained by DIDA-Net over M$^3$SDA-$\beta$ is about 7\% on both datasets and over MCD is 8.3\% on Digit-Five, which justifies the effectiveness of instance domain adaptation. Compared to the latest MSDA methods, CMSS~\cite{yang2020curriculum} doubles the model size than DIDA-Net due to the use of an auxiliary CNN model for domain and sample selection; LtC-MSDA~\cite{wang2020learning} also utilizes pseudo-labeling for estimating class prototypes for the target domain, but it consumes much more memory and computation than DIDA-Net because of maintaining a global knowledge graph (DIDA-Net only adds a small sub-network to the mainstream CNN backbone). In terms of performance, DIDA-Net beats CMSS with 3.7\% and LtC-MSDA with 2.6\% on Digit-Five. DRT~\cite{li2021dynamic} is the most related work to ours. DRT turns MSDA to single-source domain adaptation by adapting model across samples. Though DRT utilizes a dynamic model, they still embrace domain-level feature alignment loss, e.g. MCD. This significantly differs from our DIDA-Net. Our DIDA-Net also achieves 2.6\% better than DRT on Digit-Five, despite having less parameters and computation costs (see Table~\ref{tab:dida_vs_drt}).

On each specific domain of both datasets, our DIDA-Net performs favorably against other baselines. We also observe that DIDA-Net performs the best on the most challenging target domains with drastic domain shift, such as SVHN on Digit-Five and Sketch on PACS---with more than 3\% improvements over the baselines.

\textbf{DomainNet}
is the most challenging MSDA dataset at present. On DomainNet, we replace the indicator function in Eq.~(\ref{eq:Lt}) with $q(\hat{y}_i^{\mathcal{T}})$ since only a small portion of samples are selected according to $q(\hat{y}_i^{\mathcal{T}}) \geq \tau$. Our DIDA-Net surpasses other static-model based methods by a clear margin, e.g. 4.5\% over CMSS and 3.6\% over LtC-MSDA. For DRT where dynamic model is used, our DIDA-Net outperforms it by 2.2\% with even less parameters and MACs (see Table~\ref{tab:dida_vs_drt}). DRT with self-training gains better performance but still under performs our DIDA-Net.

It is worth noting that on Quickdraw and Clipart, where the visual representations are drastically different from other domains, DIDA-Net achieves the biggest improvement among all MSDA methods over the source-only baseline. The other MSDA methods~\cite{Peng_2019_ICCV,yang2020curriculum,li2021dynamic}, usually based on domain alignment, may suffer from negative transfer---the domain alignment can even harm the discriminative feature learning.  This can lead to poor class-wise separability and thus inferior performance to our DIDA-Net. 
This comparison suggests that instance domain adaptation can better handle large domain shift. Our DIDA-Net does not work very well on Infograph. This is probably caused by Infograph's images with large image areas containing irrelevant information. As a result, pseudo labels obtained on Infograph are less accurate that directly impact on the instance domain adaptation learning. This might be improved by introducing stronger regularization methods to suppress the negative impact brought by noisy pseudo labels. To sum up, compared with other methods, our DIDA-Net achieves better performance on Quickdraw and Clipart because of alleviating negative transfer (other methods may suffer from negative transfer), but worse performance on Infograph mainly due to noisy labels caused by noisy image content (other methods does not use  pseudo-labels).

\begin{table}[t]
    \caption{Ablation study on the DIDA module.} 
    \label{tab:ablation_pacs}
    \centering
    \begin{tabular}{l|c}
    \toprule
    \textbf{Method} & \textbf{Avg} \\
    \midrule
    FixMatch & 88.86 \\
     + DIDA & \textbf{91.83} \\
     + static CNN & 90.39 \\
     + DIDA w/o main stream $f_{\theta_s}$ & 80.78\\
    \bottomrule
    \end{tabular}
\end{table}

\begin{table}[t]
    \caption{DIDA vs. static CNN with different dilation rates.}
    \label{tab:ablation_dilation}
    \centering
    \begin{tabular}{l|cc}
    \toprule
    \textbf{Method}  & \textbf{DIDA} & \textbf{Static CNN}\\
    \midrule
    dilation=\{1, 2\} 	&\textbf{91.83} & 90.39\\
    dilation=1 & 90.69 & 89.91\\
    dilation=2 & 90.08 & 89.68 \\
    \bottomrule
    \end{tabular}
\end{table}

\subsection{Further Analysis}
\label{subsec:futher_analy}
In this section, we conduct further experiments on PACS to better understand our DIDA-Net.

\paragraph{Effectiveness of the DIDA module}
The DIDA module is the main contribution in this work, which transforms a static model into a dynamic model that can produce instance-adaptive features. To evaluate its effectiveness, we simply remove this module from the model, which reduces to the pseudo-labeling method based on FixMatch. We conduct this ablation study on PACS and show the results in Table~\ref{tab:ablation_pacs}. We observe that without the DIDA module, the model's performance drops 2.97\% from 91.83\% to 88.86\%.

\paragraph{Importance of the kernel generator}
To verify that the improvement is mainly brought by the dynamic design in the DIDA module (attributed to the kernel generator) rather than increasing the model capacity with more parameters, we replace the kernel generator in the DIDA module with vanilla convolutions.\footnote{When $c=512, m=16$, \#param of the DIDA module: $\sim$24.6K, \#param of the static CNN: $\sim$24.9K.} Note that such a modification changes the nature of the module and results in a multi-branch CNN model. Table~\ref{tab:ablation_pacs} shows that the DIDA module is clearly better than the static CNN, suggesting that dynamic neural networks are the key to instance domain adaptation.

\begin{table}[t]
    \caption{Ablation study on the channel reduction layer and the kernel size of dynamic kernels.} 
    \label{tab:ablation_dida_conv}
    \centering
    \begin{tabular}{l|c}
    \toprule
    \textbf{Method} & \textbf{Avg} \\
    \midrule
      standard DIDA-Net & \textbf{91.83}\\
      top branch with 1$\times$1 kernel & 90.40\\
     `conv1' not shared among three branches &90.30\\
    \bottomrule
    \end{tabular}
\end{table}

\begin{figure}[t]
    \centering
    \includegraphics[trim=10 2 30 15,clip,width=0.8\columnwidth]{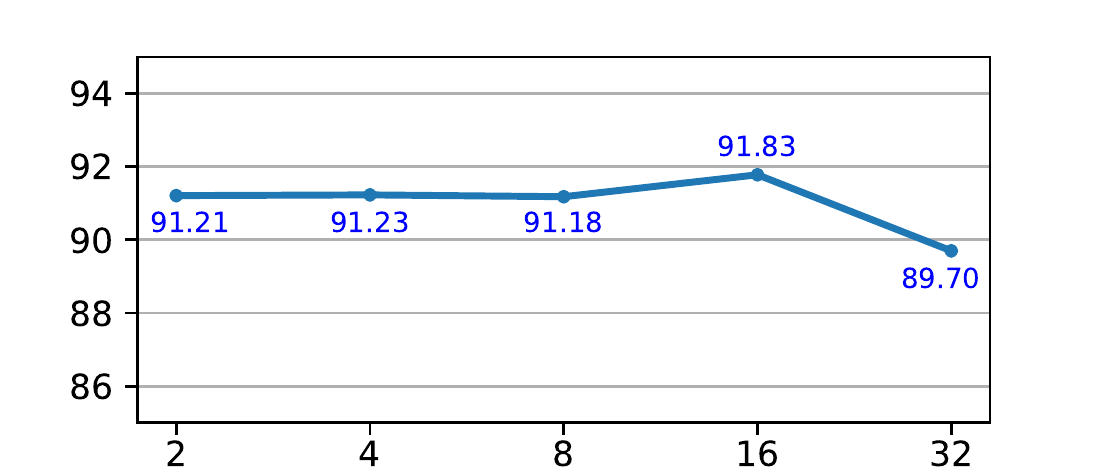}
    \caption{Accuracy of the DIDA-Net with $m$ varied.}
    \label{fig:ablation_m}
\end{figure}

\begin{table}[t]
\centering
    \caption{Analysis of where to apply the DIDA module.} 
    \label{tab:ablation_where}
    \resizebox{\columnwidth}{!}{
    \begin{tabular}{l|cccc|ccc}
    \toprule
    \textbf{Method} & \texttt{res1} & \texttt{res2} & \texttt{res3} & \texttt{res4} & \texttt{res14} & \texttt{res24} & \texttt{res34} \\
    \midrule
    \textbf{Avg}  &88.96  &90.72 &90.14  &\textbf{91.83} &90.19 &88.47 &90.05\\
    \bottomrule
    \end{tabular}
    }
\end{table}

\begin{table}[t]
    \centering
    \caption{DIDA-Net with domain alignment.} 
    \label{tab:dida_fda}
    \centering
    \begin{tabular}{l|c}
    \toprule
    \textbf{Method} & \textbf{Avg} \\
    \midrule
     DIDA-Net & \textbf{91.83} \\
     + MMD~\cite{gretton2012kernel} & 91.75 \\
     + MCD~\cite{2018Maximum} & 91.22 \\
     + DANN~\cite{ganin2016domain} & 90.58 \\
    \bottomrule
    \end{tabular}
\end{table}

\begin{table}[t]
    \caption{DIDA-Net vs. DRT~\cite{li2021dynamic}.} 
    \label{tab:dida_vs_drt}
    \centering
    \begin{tabular}{l|cc|c}
    \toprule
    \textbf{Method} & \#Param &\#MACs &\textbf{Avg} \\
    \midrule
     DIDA-Net &11.22M &1.82G & \textbf{91.83} \\
     DRT\textsuperscript{} &13.99M &2.23G & 87.30 \\
     DRT + FixMatch &13.99M &2.23G & 89.84 \\
    \bottomrule
    \end{tabular}
\end{table}

\begin{table}[t]
    \caption{DIDA-Net vs. attention-based methods.} 
    \label{tab:dida_vs_attention}
    \centering
    \begin{tabular}{l|c||c|c}
    \toprule
    \textbf{Method} & \textbf{Avg} &\textbf{Method} &\textbf{Avg} \\
    \midrule
     DIDA-Net & \textbf{91.83} & SENet~\cite{hu2018squeeze} &89.55\\
     CBAM~\cite{woo2018cbam} &86.17  &Self-attention~\cite{vaswani2017attention}& 91.27 \\
    \bottomrule
    \end{tabular}
\end{table}

\begin{table}[t]
    \caption{DIDA-Net vs. residual adapters~\cite{rebuffi2018efficient}.} 
    \label{tab:dida_vs_res_adapter}
    \centering
    \begin{tabular}{l|c}
    \toprule
    \textbf{Method} & \textbf{Avg}  \\
    \midrule
     DIDA-Net & \textbf{91.83} \\
     Residual adapters~\cite{rebuffi2018efficient} &72.93 \\
    \bottomrule
    \end{tabular}
\end{table}

\paragraph{Significance of the residual design}
The DIDA module is on a residual branch by default. 
To further evaluate the significance of this residual design, we remove the static sub-network $f_{\theta_s}$ and keep only the DIDA module. In this case, the output of DIDA module is the final feature representation. As shown in Table 2, without the residual design, the performance decreases by 10.05\%. This can be explained as follows: data samples from source and target domains are of the same labels/classes, but different in styles,  with the two parts modeled in two additive branches, i.e. the static sub-network and the DIDA module respectively. Without the static sub-network for the residual design, it is impossible to model the shared classes information across domains, thus causing the big drop for classification task.

\paragraph{Dilation rates in the kernel generator}
Recall that our dynamic kernels have multiple scales, i.e.~different dilation rates, for capturing objects of different sizes (see Fig.~\ref{fig:idam_archi}). We evaluate this design choice by comparing it with the model that has only a single dilation rate, either 1 or 2. The results are shown in the first column of Table~\ref{tab:ablation_dilation} where we can see that the multi-scale version is clearly better than the single-scale version---with more than 1\% gap.

Table~\ref{tab:ablation_dilation} also shows the comparison of the DIDA and the static CNN under the setting of different dilation rates. From the first row, we can see that the dynamic design of DIDA brings 1.44\% improvement over the static CNN. Moreover, for our DIDA, using different dilation rates (dilation=\{1,2\}) improves the accuracy by 1.14\% when compared with dilation=1. These two results show that the dynamic design contributes more to the performance improvement than the dilation rates, e.g. 1.44\% vs. 1.14\%. On the other hand, when a single dilation rate is used (the last two rows), our DIDA still outperforms its static counterpart by 0.78\% and 0.4\% respectively. This verifies that our DIDA can achieve better performance than the static ones even under the setting of single dilation.

\paragraph{Kernel sizes of the dynamic kernels}
To investigate the influence of the kernel size, we replace the top branch 3$\times$3 dynamic convolutional kernel with 1$\times$1 kernel. As shown in Table~\ref{tab:ablation_dida_conv}, 1$\times$1 kernel decreases the performance, suggesting that the 3$\times$3 convolutional kernel is more effective for capturing objects of different scales.

\paragraph{Sharing channel reduction layer in the kernel generator}
By default, we share the channel reduction layer (`conv1' in Fig.~\ref{fig:idam_archi}) among three branches to save parameters. We compare this design choice with another variant where each branch has a different channel reduction layer. From Table~\ref{tab:ablation_dida_conv}, we can see that the average accuracy of this variant is 90.30\%, worse than the shared version of standard DIDA module. This is probably because the shared `conv1' receives supervision signal from all three branches when it is updated by back-propagation, thus capturing related information from all three branches for kernel generation.

\paragraph{Where to insert the DIDA module}
We evaluate this design choice by applying the DIDA module to different layers in the CNN model. We use the ResNet architecture, which has four residual blocks (\texttt{res1-4}). For notation, \texttt{res4} means adding the DIDA module's output to \texttt{res4}'s features (both sharing the same input); \texttt{res34} means applying two DIDA modules independently to \texttt{res3} and \texttt{res4}; and so on. The results are presented in Table~\ref{tab:ablation_where}. In terms of the single-layer variants, the best result is obtained by applying the DIDA module to \texttt{res4}, which is the top layer containing the most semantic information. We then try including more layers, namely \texttt{res14}, \texttt{res24} and \texttt{res34}. It turns out that having more layers equipped with the DIDA module does not help further.

\paragraph{Sensitivity of $m$}
Fig.~\ref{fig:ablation_m} shows the results of varying the channel reduction rate $m$ from 2 to 32. It can be seen that $m=16$ gives the best result. Moreover, we observe that the performance is not too sensitive to $m$.

\begin{figure}[t]
    \centering
    \subfloat[Static model]{\label{fig:visual_wo_difc}
        \includegraphics[trim=60 60 60 65,clip,width=0.45\columnwidth]{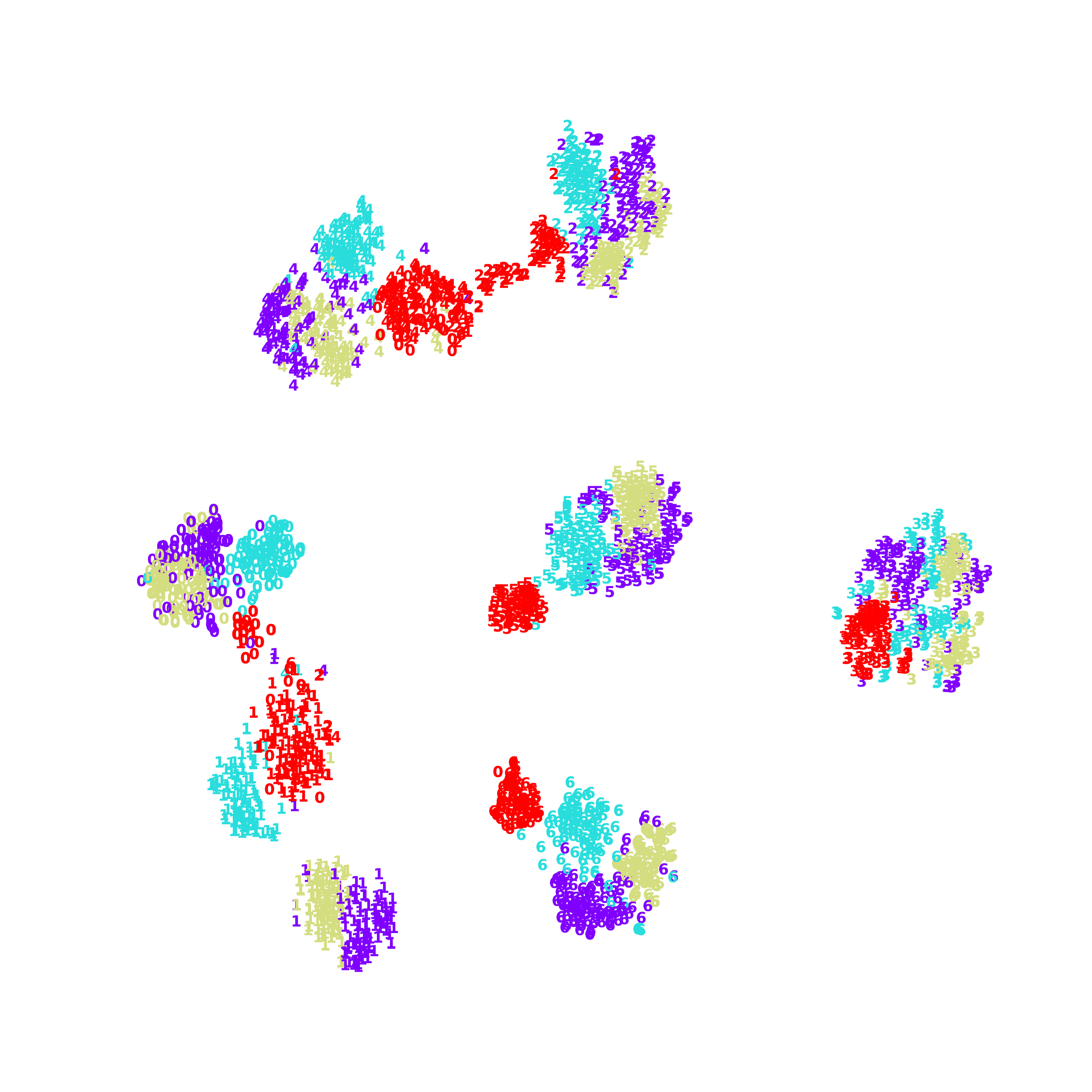}
    }
    ~\hfill
    \subfloat[DIDA-Net]{\label{fig:visual_difc}
        \includegraphics[trim=60 65 60 70,clip,width=0.45\columnwidth]{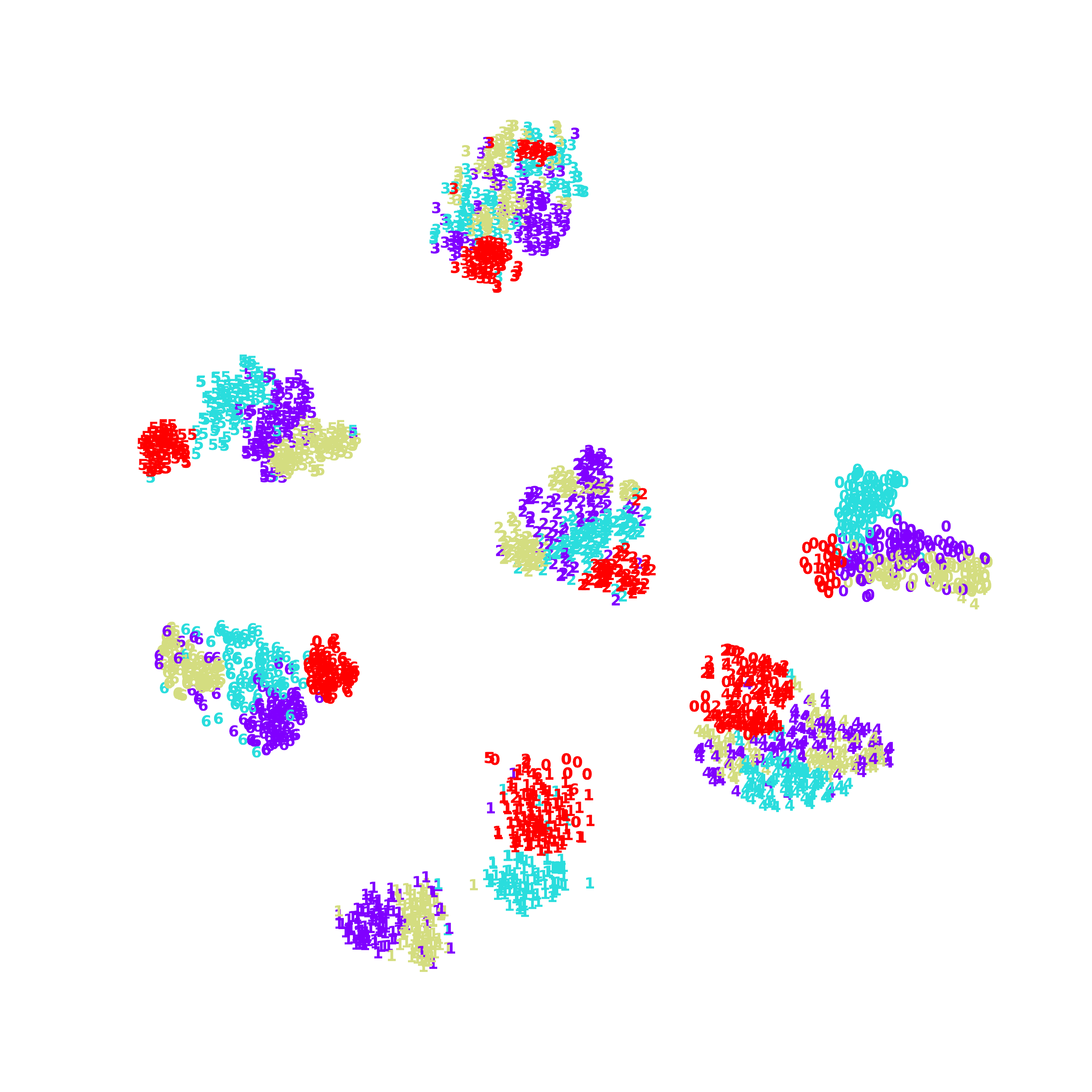}
    }
    \caption{Visualization of features from a static model (w/o DIDA module) and DIDA-Net (w/ DIDA module) on PACS using t-SNE~\cite{maaten2008visualizing}. Red color denotes the target domain (Sketch) while other colors denote the source domains. The digits 0-7 denote the 7 classes. Best viewed with zoom-in.}
    \label{fig:visualize}
\end{figure}

\paragraph{Whether domain alignment is still necessary} We introduce three popular domain-level feature alignment losses to DIDA-Net. They are metric learning-based method (MMD loss), class-wise feature alignment loss across domains (MCD), and adversarial training-based method (DANN). The results are reported in Table~\ref{tab:dida_fda}. We can see that all of these domain-level feature alignment methods fail to improve the accuracy of DIDA-Net. This is probably because our DIDA-Net has already achieved class-wise alignment by instance adaptation (see the visualization of feature distribution in Fig.~\ref{fig:visualize}), leading to feature alignment across domain redundant.

\paragraph{DIDA vs. DRT}  We further compare our DIDA-Net with the state-of-the-art  DRT~\cite{li2021dynamic}, which is also a dynamic model based domain adaptation method. From the results in Table~\ref{tab:dida_vs_drt}, we can see though our DIDA-Net has fewer model parameters and MACs (Multiply–Accumulate Operations), it still clearly surpasses DRT and DRT+FixMatch by about 4.5\% and 2.0\% accuracy on PACS. This demonstrates both the effectiveness and efficiency of our DIDA-Net. 

\paragraph{DIDA vs. attention-based methods} Attention module is another kind of dynamic models, in parallel to these methods weighted sum several kernels, e.g. DRT. Table~\ref{tab:dida_vs_attention} shows the comparison of the DIDA module with SENet~\cite{hu2018squeeze}, CBAM~\cite{woo2018cbam}, and Self-attention~\cite{vaswani2017attention}. It is clear that the DIDA-Net is better than all these attention-based methods, probably because dynamic kernels of different dilation rates in the DIDA module can better capture the objects of different scales than simply re-scaling features.

\paragraph{DIDA vs. residual adapter} Similar to parallel residual adapters~\cite{rebuffi2018efficient}, the DIDA-Net uses residual branch as an adapter. We show in Table~\ref{tab:dida_vs_res_adapter} that the DIDA-Net outperforms the parallel residual adapter by a large margin. We attribute the superior performance of DIDA-Net to the instance adaptation scheme and its robustness to scale variation of objects.

\begin{table}[H]
    \caption{The complexity of DIDA-Net and other UDA methods. }
    \label{tab:complexity}
    \centering
    \begin{tabular}{l|cc}
    \toprule
    \textbf{Method} & \textbf{\#Param} &\#MACs\\
    \midrule
     DIDA-Net &11.22M &1.82G\\
     DRT~\cite{li2021dynamic} &13.99M &2.23G  \\
     M$^3$SDA-$\beta$~\cite{Peng_2019_ICCV} & 11.00M &1.82G \\
     MCD~\cite{2018Maximum} & 11.00M &1.82G \\
     MMD~\cite{gretton2012kernel} &11.00M &1.82G  \\
     DANN~\cite{ganin2016domain} &11.00M &1.82G  \\
    \bottomrule
    \end{tabular}
\end{table}

\begin{table}[t]
    \caption{The accuracy of DIDA-Net on a semi-supervised learning dataset, CIFAR-10. The averaged accuracy and its standard derivation of three runs are reported. * denotes our implementation.}
    \label{tab:dida_ssl}
    \centering
    \begin{tabular}{l|ccc}
    \toprule
    \textbf{Method}  & 40 labels & 250 labels & 4000 labels\\
    \midrule
    FixMatch \textsuperscript{*} &69.88$\pm$7.88 &88.27$\pm$0.52 &90.16$\pm$0.26\\
    + DIDA &78.98$\pm$2.47  &88.37$\pm$0.33 &91.18$\pm$0.10\\
    \bottomrule
    \end{tabular}
\end{table}

\paragraph{Computational complexity} In Table~\ref{tab:complexity}, we compare the computational complexity of DIDA-Net with DRT~\cite{li2021dynamic}, M$^3$SDA-$\beta$~\cite{Peng_2019_ICCV}, MCD~\cite{2018Maximum}, DANN~\cite{ganin2016domain} and MMD~\cite{gretton2012kernel}. Except DIDA-Net and DRT, the backbone of the other methods is a static ResNet-18 without any dynamic design (here, we do not consider the classifier). We can see that our DIDA-Net only introduces about 0.22M extra parameters to the vanilla ResNet-18 but keeps the MACs (Multiply–Accumulate Operations) the same. In contrast, DRT increases 2.99M extra parameters and 0.41G MACs. Overall, our DIDA-Net achieves better performance than the other UDA methods (see Table~\ref{tab:compare_stateoftheart}), but does not significantly increase the computational complexity.

\subsection{Extend DIDA to Semi-Supervised Learning}
Due to instance-adaptive parameters, our DIDA does not need to explicitly model the domain discrepancy. This presumably makes the DIDA applicable for semi-supervised learning tasks. Therefore, we further apply our DIDA-Net to semi-supervised learning task by conducting experiments on  CIFAR-10~\cite{krizhevsky2009learning}. In Table~\ref{tab:dida_ssl}, we compare our DIDA-Net with the baseline, i.e., FixMatch~\cite{sohn2020fixmatch}. We can see that our DIDA-Net can improve the performance over the FixMatch baseline, with 9.1\% performance gain for 40 labels (4 labels per class) and 1.02\% for 4000 labels. This is another exciting advantage of our method.

\section{Visualization}
\label{sec:visualization}
We visualize the final features ($Z_i$ in Eq.~\eqref{eq:difc}) for DIDA-Net and the static model in Fig.~\ref{fig:visualize} to better understand the effect of instance domain adaptation. By comparing Fig.~\ref{fig:visualize}(a) with (b), we observe that 1) DIDA-Net's features are better aligned across different domains (including source and target) compared to the baseline's features, and 2) different classes for the target domain (red-colored) in DIDA-Net's feature space are much more separable. This verifies that our DIDA-Net can implicitly achieve class-wise alignment.

\section{Conclusion}

In this paper, we argued that UDA datasets' data distributions are often inconsistent with the existing domain definitions, which cannot be solved by applying domain alignment to static models. Therefore, we introduced the idea of viewing each instance as a fine domain and proposed to use dynamic neural networks to facilitate the learning of a common feature space between source and target. Our model, DIDA-Net, with a simple training strategy, was validated on two single-source domain adaptation and three MSDA datasets with performance superior to current state of the art.



\bibliographystyle{IEEEtran}

\bibliography{dida.bib}

\vfill

\end{document}